\documentclass[sigconf,nonacm]{acmart}

\setcopyright{none}
\AtBeginDocument{%
  }

\usepackage{pifont}
\usepackage{multirow}
\usepackage{colortbl}
\usepackage{enumitem}
\usepackage{threeparttable}
\usepackage[most]{tcolorbox}
\usepackage{tabularx}
\usepackage{booktabs}
\usepackage{graphicx}
\usepackage[table]{xcolor}
\usepackage{wasysym}
\usepackage{array}
\usepackage{makecell}
\usepackage{svg}
\usepackage[normalem]{ulem}

\newcolumntype{Y}{>{\centering\arraybackslash}X}

\newtcolorbox{promptbox}[1]{
    breakable,
    title=#1,
    colback=white,
    colframe=black!70,
    fonttitle=\bfseries,
    enhanced,
    before skip=10pt,
    after skip=10pt
}

\definecolor{blue-violet1}{rgb}{0.20, 0.20, 0.60}

\newcommand{\yxnew}[1]{\textcolor{black}{#1}}
\definecolor{yxgreen}{RGB}{0,160,70}

\definecolor{blue-violet}{rgb}{0.54, 0.17, 0.89}

\begin{document}

\title{Towards Root Memory: Benchmarking and Enhancing Implicit Logical Memory Retrieval for Personalized LLMs}

\author{Hongxun Ding}
\authornote{Equal contribution.}
\affiliation{%
  \institution{University of Science and Technology of China}
  \country{China}
}
\email{hongxunding02@gmail.com}

\author{Xiang Yu}
\authornotemark[1]
\affiliation{%
  \institution{University of Science and Technology of China}
  \country{China}
}
\email{yux661988@gmail.com}

\author{Chengbing Wang}
\affiliation{%
  \institution{University of Science and Technology of China}
  \country{China}
}
\email{wwq197297@mail.ustc.edu.cn}

\author{Jianfei Xiao}
\affiliation{%
  \institution{University of Science and Technology of China}
  \country{China}
}
\email{jianfeixiao@mail.ustc.edu.cn}

\author{Keqin Bao}
\affiliation{%
  \institution{University of Science and Technology of China}
  \country{China}
}
\email{baokq@mail.ustc.edu.cn}

\author{Wenjie Wang}
\affiliation{%
  \institution{University of Science and Technology of China}
  \country{China}
}
\email{wenjiewang96@gmail.com}

\author{Xiangnan He}
\authornote{Corresponding author.}
\affiliation{%
  \institution{University of Science and Technology of China}
  \country{China}
}
\email{xiangnanhe@gmail.com}

\renewcommand{\shortauthors}{Ding et al.}

\begin{abstract}
Memory systems are essential for personalized Large Language Models (LLMs). However, retrieval within these systems primarily relies on semantic similarity, potentially missing logically critical memories with limited semantic overlap. We refer to this failure as \textbf{semantic deflection}, which remains insufficiently evaluated by existing benchmarks. To address this gap, we construct \textbf{IMLogic}, a benchmark for implicit logical memory retrieval in long dialogues. Each question is designed around the divergence between semantic and logical relevance, with diagnostic response options that probe semantic deflection, hallucination, and lack of personalization. To help systems utilize associations that are difficult to discover through semantic matching, we introduce \textbf{root memory}, a structured representation that distills reusable, response-guiding logic from user histories through LLM reasoning and grounds it in supporting evidence. We further develop \textbf{RootMem}, a plug-and-play framework that constructs and routes root memories to complement semantic retrieval. Extensive experiments reveal persistent challenges across retrieval methods and memory systems, including substantial preference-driven errors even under full-memory access. RootMem consistently improves performance, demonstrating the potential of structured personalized logic while leaving considerable challenges unresolved. Our benchmark and code are available at \url{https://github.com/ThisIsCosine/IMLogic}.
\end{abstract}

\keywords{Large Language Models, Personalization, Implicit Logical Memory Retrieval}
\maketitle

\section{Introduction}

As Large Language Models (LLMs) are increasingly deployed in user-centric applications~\citep{zhao2025nextquill,wang2025think,qiu2025latent,qiu2025measuring,wang2026perm,zhang2025reinforced,xu2025personalized}, memory systems have become essential for sustained personalized interactions~\citep{intro1zhang2025survey, intro1wu2025human, hu2026memoryageaiagents,xiao2026alpsbench,zhang2026explicit,zhao2026don}. These systems accumulate extensive personal data, behavioral patterns, and preferences as user memories over time~\citep{intro2tan2025prospect, intro2zhong2024memorybank}. To generate accurate and personalized responses, memory systems often need to extract relevant memories from previous context, then store and manage them for subsequent retrieval in response to new queries.

\begin{figure}[t]
    \centering
    \includegraphics[
        width=1.05\columnwidth,trim=11.96cm 1.6cm 6.46cm 0cm,clip
    ]{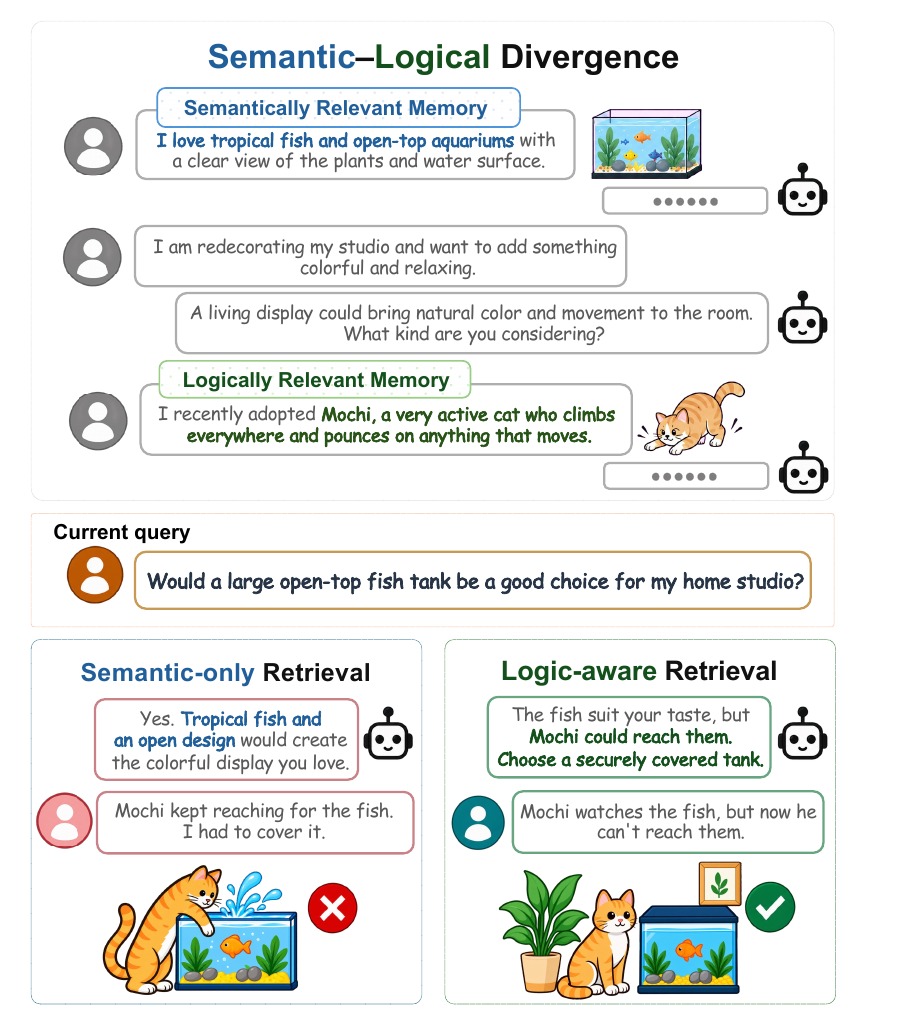}
    \caption{A dialogue example illustrating divergence between semantic and logical relevance of user memories to a query.}

    \label{fig:intro_semantic_deflection}
\end{figure}

Existing retrieval methods employed in memory systems largely rely on semantic similarity~\citep{hu2026memoryageaiagents}. However, semantic similarity does not always reflect the logical relevance of memories to the current query (e.g., through constraints, motives, or conditions).
As shown in Figure~\ref{fig:intro_semantic_deflection}, the user's preference for fish tanks is semantically close to the query, but an appropriate response also needs to consider the semantically distant habits of the user's cat: the cat may harm fish in an open-top tank. This example illustrates the divergence between semantic and logical relevance. We refer to the resulting retrieval failure as \textbf{semantic deflection}: existing methods may fail to recall logically relevant but semantically distant information, leading LLMs to respond to superficial intent and ultimately undermining trust and user experience. To address this challenge, we study \textbf{implicit logical memory retrieval}: discovering and utilizing memories from user history that are weakly related to the current query semantically but logically critical for a personalized response.

Despite its importance for LLM personalization, existing memory benchmarks remain insufficient for systematically evaluating semantic deflection.
Existing benchmarks primarily focus on locating historical information in complex contexts~\citep{maharana2024locomo, wulongmemeval} or on reasoning over implicit clues within interactions~\citep{jiang2025personamem,li2025implexconv,li2026locomoplus}.
To effectively evaluate semantic deflection, a benchmark should not only cover complex settings such as long conversations and multiple memory entries, but also explicitly construct evaluation scenarios in which semantic and logical relevance diverge.


To address this gap, we introduce \textbf{IMLogic}, a benchmark for implicit logical memory retrieval in long conversations, built from HaluMem’s~\citep{chen2025halumem} interaction histories and memory entries. An LLM mines candidate memory pairs with logical associations, which are then verified by experts. A \textit{Generator-Judger-Refiner} pipeline constructs QA pairs whose queries overlap semantically with the \textbf{semantic memory}, while correct responses depend on the semantically distant \textbf{logical memory}. Each question includes a correct response and three diagnostic alternatives targeting semantic deflection, hallucination, and lack of personalization. Experts screen the generated samples, followed by independent human review of a random sample to assess quality. IMLogic comprises approximately 15,000 user memories and 2,216 implicit logical QA pairs, supporting both memory-level evaluation with memory entries and conversation-level evaluation with raw dialogues.

Building on this benchmark, we find that mainstream memory systems with different retrieval strategies struggle with implicit logical memory retrieval; in the conversation-level evaluation, even the best-performing memory system baseline achieves only 44.18\% accuracy.
To help systems utilize associations that are difficult to discover through semantic matching, we introduce \textbf{root memory}, a structured memory representation of reusable personalized logic distilled from user history through LLM reasoning to guide responses.
In this way, memory systems capture not only what information is stored, but also how it should inform future personalized responses.

To instantiate this concept, we propose \textbf{RootMem}, a plug-and-play framework that constructs and utilizes root memories through LLM-based reasoning. During offline construction, RootMem organizes historical memories into structured root memory units that preserve response-guiding logic and its personalized grounding. At inference time, an LLM-based router identifies the units relevant to the current query and combines them with conventional semantic retrieval, enabling the memory system to incorporate logical relevance beyond semantic matching.

We conduct extensive experiments on IMLogic to systematically evaluate diverse retrieval methods and memory systems. The results reveal persistent challenges across retrieval methods and end-to-end systems, with substantial preference-driven errors even under full memory access. RootMem achieves a 27.23\% relative improvement over the strongest retrieval baseline and 
an average 26.36\% relative gain in our memory system integration experiments.
Nevertheless, substantial errors remain, particularly in open-ended generation, highlighting both the potential of structured personalized logic and the challenges ahead.

Overall, our contributions are fourfold:

\begin{itemize}[leftmargin=*, nosep]
    \item We characterize semantic--logical divergence in personalized memory retrieval and the associated failure, semantic deflection, and formulate the problem of implicit logical memory retrieval.
    \item We construct IMLogic from existing long-term user memories to evaluate semantic deflection, with diagnostic response options and evaluation at both memory and conversation levels.
    \item We introduce root memory, a structured representation of reusable personalized logic grounded in supporting evidence, and develop a plug-and-play framework for its construction and routing.
    \item Through extensive experiments, we reveal persistent limitations of existing retrieval methods and memory systems, demonstrating both the potential and remaining challenges of structured personalized logic organization.
\end{itemize}

\section{Related Work}

As summarized in Table~\ref{tab:benchmark_compare}, existing benchmarks evaluate preference adherence (PrefEval~\citep{zhao2025prefeval}), long-context recall (LoCoMo~\citep{maharana2024locomo}, LongMemEval~\citep{wulongmemeval}), persona modeling (PersonaMem~\citep{jiangknow}), and memory processing (HaluMem~\citep{chen2025halumem}, AlpsBench~\citep{xiao2026alpsbench}). PersonaMem-v2~\citep{jiang2025personamem} studies implicit persona inference, ImplexConv~\citep{li2025implexconv} evaluates implicit reasoning in multi-session conversations, and LoCoMo-Plus~\citep{li2026locomoplus} assesses responses against latent contextual constraints. IMLogic centers on diagnosing semantic deflection in personalized responses. It mines logical associations from existing user histories and provides diagnostic response options to distinguish semantic deflection, hallucination, and lack of personalization. It also supports multi-level evaluation.

\begin{table}[h]
\centering
\small
\caption{Comparison of IMLogic with representative memory benchmarks across evaluation design and context scale. Column definitions and support levels are given below.}
\label{tab:benchmark_compare}


\begin{threeparttable}
\scriptsize
\renewcommand{\arraystretch}{1.15}
\setlength{\tabcolsep}{1.8pt}

\begin{tabularx}{\columnwidth}{
>{\raggedright\arraybackslash}p{0.20\columnwidth}
*{5}{Y}
>{\centering\arraybackslash}p{0.15\columnwidth}
>{\centering\arraybackslash}p{0.12\columnwidth}
}
\toprule

\multirow{2}{*}{\textbf{Benchmark}} &
\multicolumn{5}{c}{\textbf{Benchmark Characteristics}} &
\multicolumn{2}{c}{\textbf{Context Statistics}} \\

\cmidrule(lr){2-6}
\cmidrule(lr){7-8}

&
\textbf{SD} &
\textbf{HD} &
\textbf{CC} &
\textbf{ME} &
\textbf{DA} &
\textbf{Conv. Len.} &
\textbf{Queries} \\

\specialrule{\lightrulewidth}{1pt}{1pt}


PrefEval~\citep{zhao2025prefeval}
& \Circle
& \Circle
& \Circle
& \Circle
& \CIRCLE
& 3K,23K,100K
& 3,000 \\

LoCoMo~\citep{maharana2024locomo}
& \Circle
& \CIRCLE
& \CIRCLE
& \CIRCLE
& \Circle
& 16K
& 1,986 \\

LongMemEval~\citep{wulongmemeval}
& \Circle
& \Circle
& \LEFTcircle
& \CIRCLE
& \Circle
& 115K,1.5M
& 500 \\

PersonaMem~\citep{jiangknow}
& \Circle
& \CIRCLE
& \CIRCLE
& \Circle
& \LEFTcircle
& 32K,128K,1M
& 5,990 \\

HaluMem~\citep{chen2025halumem}
& \Circle
& \CIRCLE
& \CIRCLE
& \CIRCLE
& \CIRCLE
& 160K,1M
& 3,467 \\

AlpsBench~\citep{xiao2026alpsbench}
& \Circle
& \CIRCLE
& \CIRCLE
& \CIRCLE
& \CIRCLE
& 5.4K
& 2,500 \\

%

\mbox{PersonaMem-v2}~\citep{jiang2025personamem}
& \Circle
& \CIRCLE
& \CIRCLE
& \Circle
& \LEFTcircle
& 32K,128K
& 5,000 \\

ImplexConv~\citep{li2025implexconv}
& \LEFTcircle
& \Circle
& \Circle
& \CIRCLE
& \Circle
& 60K
& 8,517 \\

LoCoMo-Plus~\citep{li2026locomoplus}
& \LEFTcircle
& \Circle
& \CIRCLE
& \Circle
& \Circle
& 16K
& 401 \\

\midrule
\textbf{IMLogic}
& \CIRCLE
& \CIRCLE
& \CIRCLE
& \CIRCLE
& \CIRCLE
& \textbf{160K}
& \textbf{2,216} \\

\bottomrule
\end{tabularx}

\begin{tablenotes}[flushleft]\scriptsize

\item \CIRCLE/\LEFTcircle/\Circle denote full/partial/no support.

\item
\textbf{SD}: Semantic Deflection Diagnosis, diagnoses failures from prioritizing semantic over logical relevance.
\textbf{HD}: History-Derived Evaluation, derives queries from established conversational histories or associated memories without constructing query-specific evidence histories;
\textbf{CC}: Conversational Coherence, maintains coherent conversational histories;
\textbf{ME}: Multi-level Evaluation, supports evaluation at multiple stages or levels beyond end-to-end answering;
\textbf{DA}: Diagnostic Analysis, provides diagnostic annotations, response options, or evaluation rules for distinguishing error types.

\item 
\textbf{Conv. Len.} reports the conversation-length configurations provided by each benchmark; multiple values denote distinct configurations rather than a continuous range.
\textbf{Queries} denotes the total number of evaluation query instances.
\end{tablenotes}

\end{threeparttable}
\end{table}

To extract relevant knowledge from large memory banks, recent LLM agents employ lexical~\citep{lexical1,lexical2,lexical3}, semantic~\citep{semantic1,fang2025lightmem}, graph-based~\citep{graph1,graph2,xu2025amem}, and hybrid retrieval~\citep{hybrid1,hybrid2,hybrid3}, often enhanced by query reconstruction (e.g., decomposition or rewriting)~\citep{hu2026evermemos}, re-ranking~\citep{rerank1,hu2026evermemos,rerank2}, or compression~\citep{compress1,compress2,compress3} to reduce noise. T-Mem~\citep{guo2026t} further incorporates future-oriented triggers to support associative memory retrieval. However, effectively utilizing logically essential memories with limited semantic overlap remains challenging. We propose RootMem, which constructs root memory units from historical user memories and routes relevant ones during inference to augment semantic retrieval with decision-guiding logic and personalized evidence.

\begin{figure*}[t]
    \centering
    \includegraphics[width=0.95\textwidth,trim=0 1.2cm 0 1cm,
        clip]{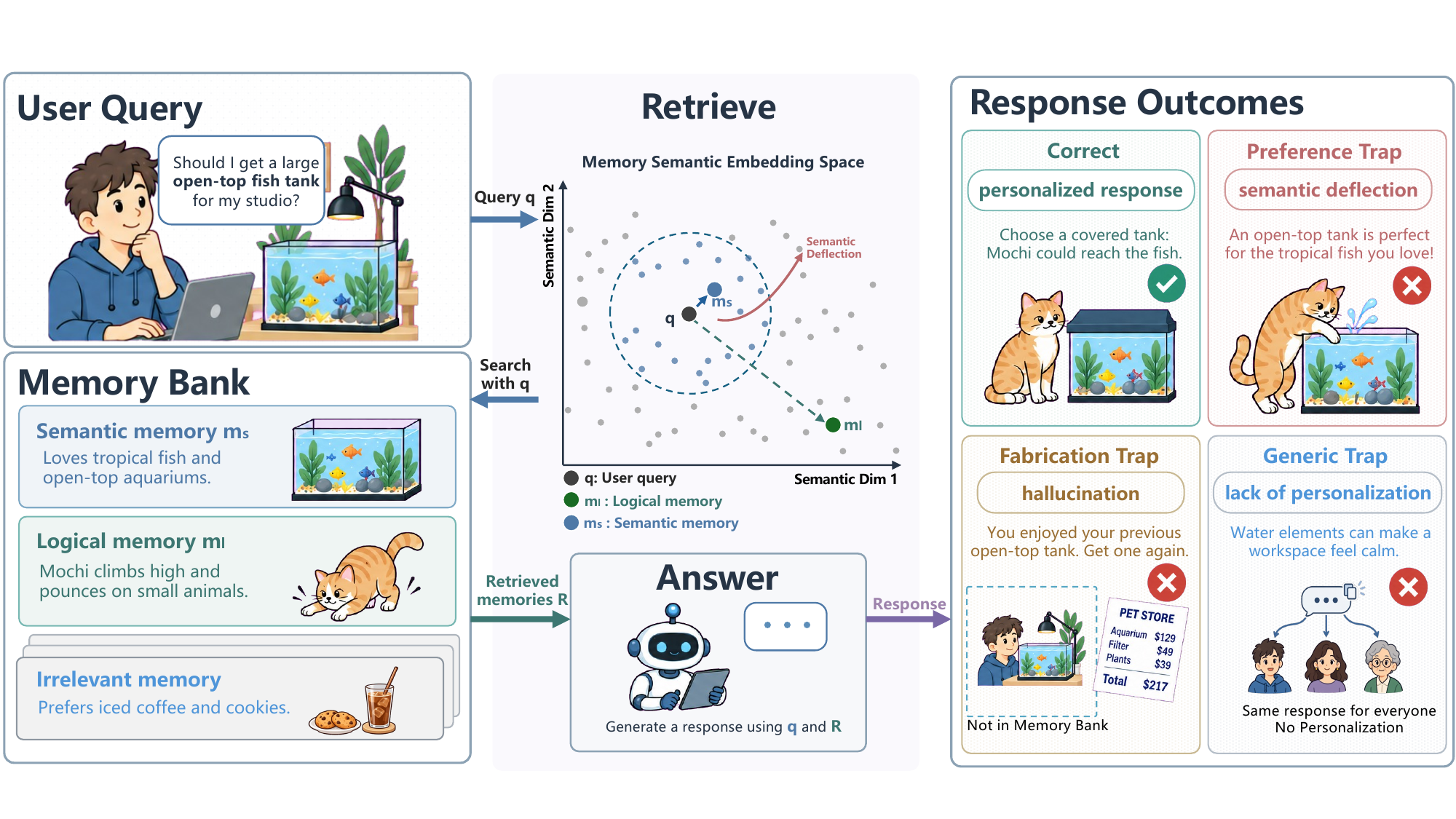}
    \caption{Evaluation workflow in IMLogic. Given a query and a memory bank, a system retrieves memories and uses them to produce a personalized response. Correct responses incorporate the logic of $m_l$, while the three traps probe semantic deflection (Preference Trap), hallucination (Fabrication Trap), and lack of personalization (Generic Trap).}
    \label{fig:response_design}
\end{figure*}

\section{The IMLogic Benchmark: Towards Implicit Logical Memory Retrieval}

%

%
%
%
\subsection{Problem Formulation}

We formalize the problem setting of \textbf{implicit logical memory retrieval} as follows. Here, implicit indicates that the query provides limited semantic cues for identifying the logically relevant memory. Given a user query $q$ and a memory bank $\mathcal{M} = \{m_1, m_2, \dots, m_N\}$, where $N$ is the total number of memory entries, we focus on a pair of memories with an implicit logical relationship, denoted as the \textbf{semantic memory} $m_s$ and the \textbf{logical memory} $m_l$. The semantic memory $m_s$ exhibits relatively high semantic overlap with $q$, but alone is insufficient to support the most appropriate personalized response. In contrast, the logical memory $m_l$ is semantically distant from $q$ but contains information that is logically important for such a response. The task is therefore to overcome semantic deflection induced by $m_s$, retrieve $m_l$, and use its underlying logic to generate accurate, personalized responses.

\subsection{Benchmark Design}

To evaluate the proposed problem setting, we introduce \textbf{IMLogic}, a benchmark for implicit logical memory retrieval. The benchmark comprises 20 long conversations averaging approximately 160K tokens each, approximately 15,000 memory entries, and 2,216 high-quality QA pairs. Its design explicitly contrasts semantic and logical relevance, derives evaluation scenarios from existing histories while preserving conversational coherence, and supports both multiple evaluation levels and diagnostic response options. Table~\ref{tab:benchmark_compare} compares these characteristics and context statistics with those of representative memory benchmarks.

\textbf{Evaluation Settings.} Figure~\ref{fig:response_design} illustrates the evaluation workflow: a system retrieves relevant evidence from user history and uses it to answer the query. We provide two evaluation modes: \textbf{memory-level} evaluation uses pre-annotated memory entries to assess logical memory retrieval without confounding effects from memory extraction, while \textbf{conversation-level} evaluation uses raw dialogues to assess the complete memory system end to end in realistic interaction settings. 
Queries cover everyday scenarios requiring personalized responses, such as seeking recommendations and engaging in casual conversations.

\textbf{Response Design.} Each query is paired with a correct response and three diagnostic alternatives: 1) \textbf{Correct} uses the logic of $m_l$ to provide an appropriate personalized response, regardless of whether $m_s$ is used; 2) 
\textbf{Preference Trap} follows the apparent preference in $m_s$ while overlooking the critical logic in $m_l$, capturing the response-level manifestation of semantic deflection; 3) \textbf{Fabrication Trap} introduces unsupported personal details to probe hallucination; and 4) \textbf{Generic Trap} provides a non-personalized response without relevant user context.

These options support multiple-choice evaluation and analysis of response-level failure modes. In open-ended evaluation, an LLM Judger uses the correct response as a reference to assess whether the generated answer incorporates the underlying logic of $m_l$.

\textbf{Semantic--Logical Divergence.} Each query is associated with memories or dialogue segments that serve as evidence for its answer. We use \textit{all-MiniLM-L6-v2}~\citep{all-MiniLM-L6-v2} embeddings to compute mean query--evidence cosine similarity, using the evidence formats available in each benchmark. For IMLogic, we additionally measure query similarity to semantic memory $m_s$, which provides relevant context but is insufficient on its own, to characterize its potential to distract retrieval from the critical logical memory $m_l$.

\begin{figure}[t]
    \centering
    \includegraphics[width=\columnwidth,
        trim=4.8cm 3.8cm 4.3cm 3.2cm,
        clip]{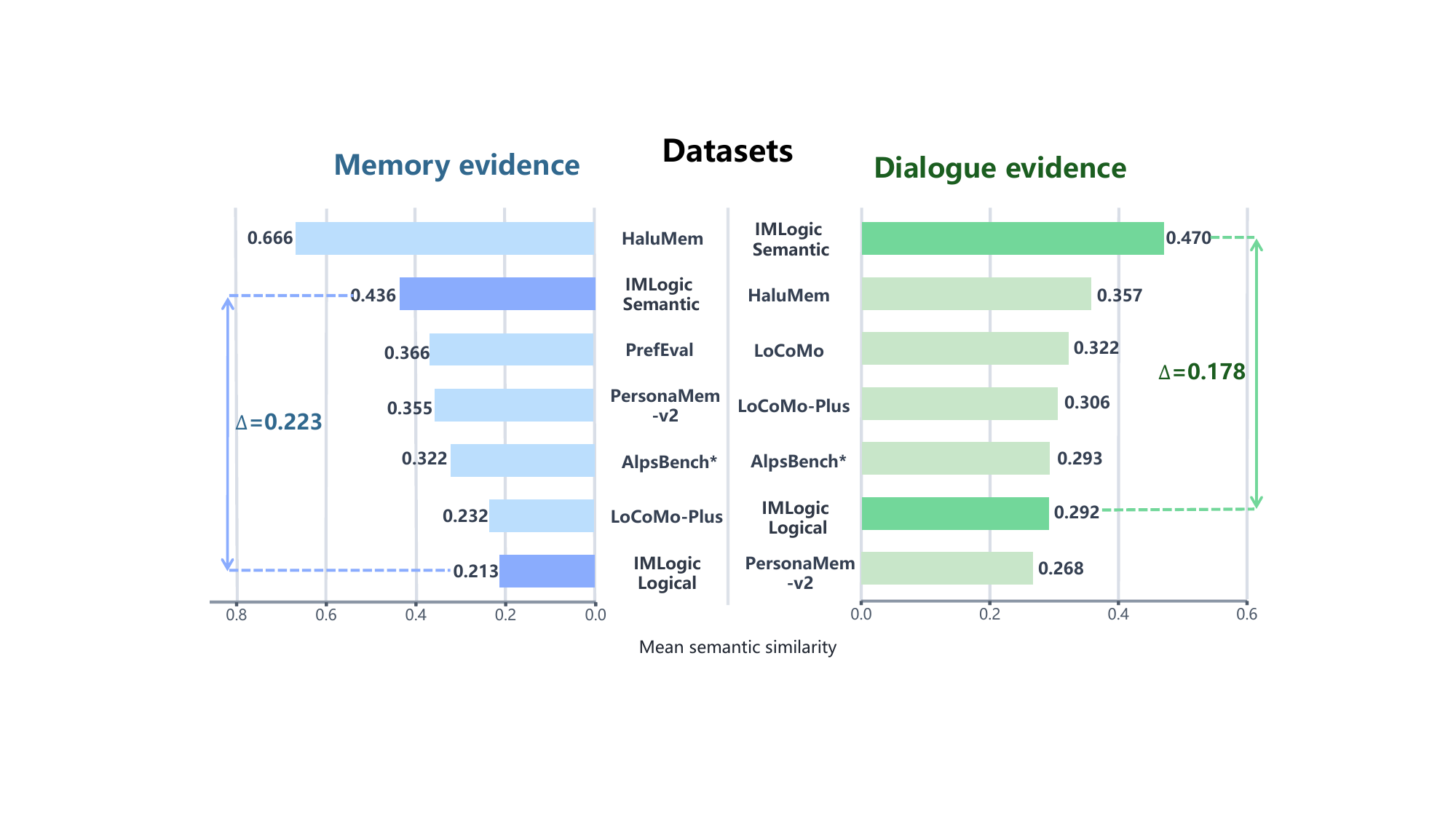}
    {\raggedright\scriptsize \textsuperscript{*}AlpsBench results use Task 4, which evaluates personalized information utilization.}\par
    \caption{Mean cosine similarity between queries and memory entries (left) or dialogue segments (right). IMLogic Semantic and IMLogic Logical denote semantic memory and logical memory, respectively.}
    \label{fig:query_evidence_similarity}

    \Description{Two horizontal bar charts compare average query--evidence cosine similarity. IMLogic semantic evidence has higher similarity than its logical evidence at both the memory entry and dialogue levels. The AlpsBench results use its utilization task, Task 4.}
\end{figure}

As shown in Figure~\ref{fig:query_evidence_similarity}, IMLogic's logical evidence has low query similarity at both levels: 0.213 for memory entries and 0.292 for dialogue evidence. Notably, the semantic–logical similarity gaps are 0.223 and 0.178, respectively, making IMLogic well suited for evaluating semantic deflection.

\begin{figure*}[t]
    \centering
    \includegraphics[width=\textwidth,
        trim=0cm 1.5cm 0cm 1.0cm,
        clip]{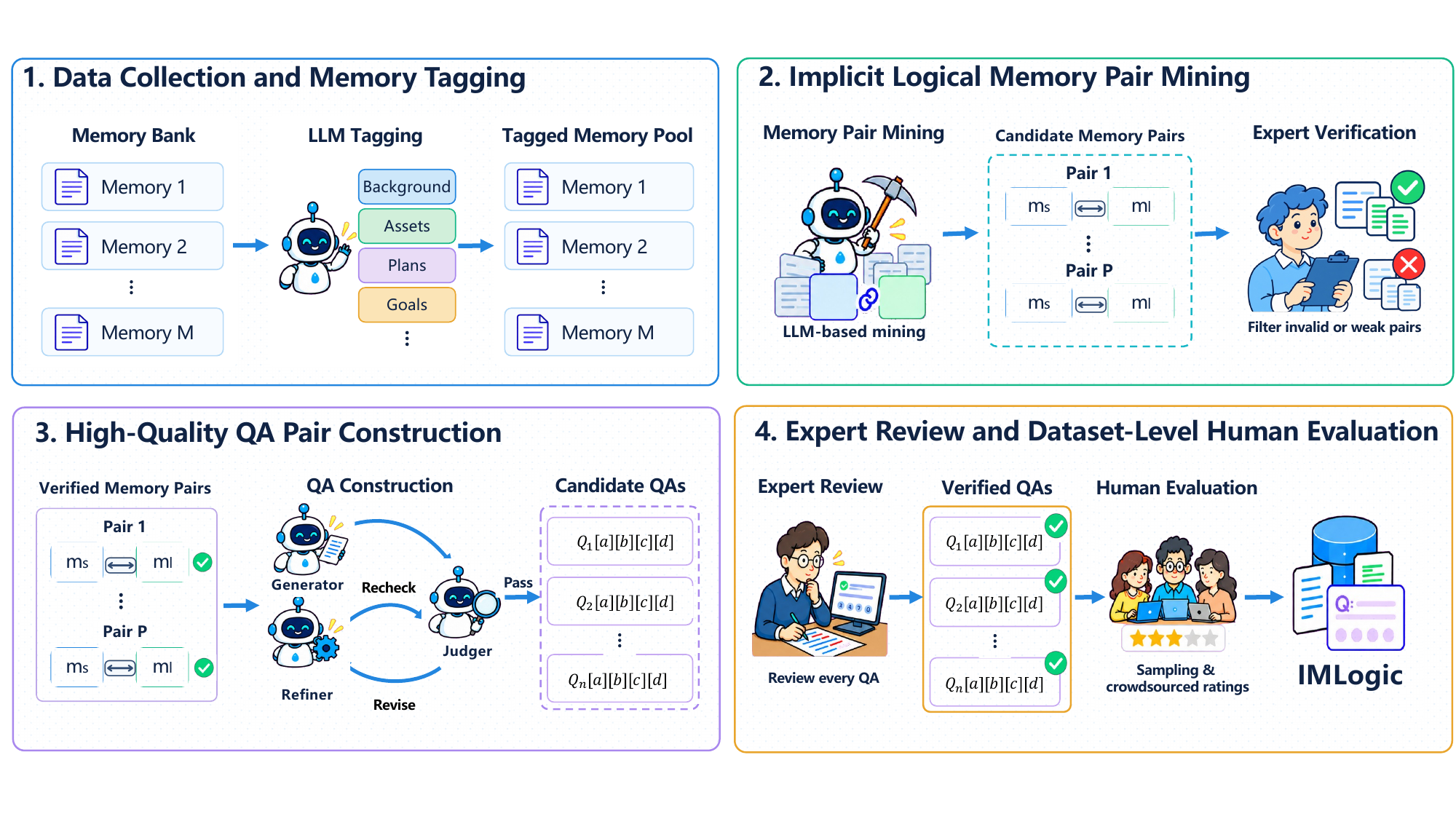}
    \caption{IMLogic construction pipeline. Existing memories are tagged to support logical association mining across categories. Expert-verified memory pairs guide QA construction through a Generator--Judger--Refiner loop. Generated instances undergo expert screening, followed by independent human evaluation of sampled retained instances.}
    \Description{Four stages show memory collection and tagging, logical pair mining with expert verification, iterative QA generation and refinement, and final expert review with independent human evaluation of a sample.}
    \label{fig:dataset}
\end{figure*}

\subsection{Benchmark Construction}
\label{sec:benchmark_construction}

Figure~\ref{fig:dataset} illustrates the IMLogic construction pipeline, which derives QA instances from logical associations within existing user histories. The construction prompts are provided in our repository.

\textbf{Data Collection and Memory Tagging.} To ground evaluation queries in pre-existing user histories, IMLogic reuses the Medium split of HaluMem~\citep{chen2025halumem}, mining logical associations from its existing memories while preserving the original dialogue histories. Subsequently, we conduct fine-grained categorical tagging on the memory bank. Utilizing an LLM, we classify the memories into nine major categories: personal background, assets, past experiences, states, opinions, goals, plans, social relationships, and others, to support subsequent pair mining.

\textbf{Implicit Logical Memory Pair Mining.}
To capture diverse logical associations across different types of personal information, we use Gemini-3-Pro~\citep{deepmind2025gemini3pro} to identify candidate memory pairs with implicit logical associations. We design a specific meta-prompt that guides the LLM to examine user memories and extract logically associated memory pairs, without imposing predefined logic categories or fixed rule templates. We then filter repetitive candidates and have experts assess the quality and logical validity of the remaining pairs before QA generation. The semantic and logical roles of the paired memories are defined relative to the generated query. We subsequently summarize the mined associations into seven logic categories through post hoc analysis, as outlined in Appendix ~\ref{sec:logic}.

\textbf{High-Quality QA Pair Construction.} Based on the filtered memory pairs, we execute a collaborative ``Generator-Judger-Refiner'' QA construction pipeline, using Gemini-3-Pro for all three agents. For the generation of the query $q$, we ensure that $q$ has high semantic overlap with $m_s$, yet correctly answering it relies on the logical constraints from $m_l$. After the Generator creates the initial QA, the Judger reviews the generated QA pair based on five core dimensions: logical rationality, semantic stealthiness, tone naturalness, option unbiasedness, and the logical superiority of the correct answer over other options. The Refiner revises instances that fail these checks using the Judger's feedback. This loop continues until the instance passes or is discarded after repeated unsuccessful revisions.


\subsection{Quality Control}
\label{sec:dataset_quality}

We apply a four-stage protocol combining model-based screening with human validation.

\textbf{(1) Pair Verification.} Experts verify the logical validity of candidate memory pairs before QA generation.

\textbf{(2) Judger--Refiner Filtering.} Generated QA pairs undergo the iterative checks described above, yielding 4,417 candidates for expert review.

\textbf{(3) Expert Filtering.} Two experts from the author team independently assess candidates for appropriateness, relevance, and personalized response quality. Table~\ref{tab:expert_filtering} reports high expert agreement as measured by Cohen's $\kappa$~\citep{cohen1960agreement}, supporting consistent screening decisions.
\begin{table}[t]
    \centering
    \small
    \scriptsize
    \setlength{\tabcolsep}{2.2pt}
    \renewcommand{\arraystretch}{0.95}
    \caption{Expert screening: sample retention and annotation agreement.}
    \label{tab:expert_filtering}
    \resizebox{\columnwidth}{!}{%
    \begin{tabular}{ccccc}
        \toprule
        \shortstack{LLM-filtered\\candidates} &
        \shortstack{Retained\\instances} &
        \shortstack{Retention\\rate} &
        \shortstack{Expert raw\\agreement} &
        \shortstack{Cohen's\\$\kappa$} \\
        \midrule
        4,417 & 2,216 & 50.17\% & 86.7\% & 0.73 \\
        \bottomrule
    \end{tabular}%
    }
\end{table}

\textbf{(4) Independent Human Re-evaluation.} Twenty CloudResearch Connect\footnote{\url{https://www.cloudresearch.com/products/participant-recruitment/}} annotators re-evaluate 30 randomly sampled instances from different users using four-point ratings. Table~\ref{tab:human-reevaluation} shows that high positive rates and raw agreement after binarization provide independent support for IMLogic's data quality. Appendix~\ref{sec:human_evaluation} provides the annotation criteria, participant demographics, and evaluation protocol, together with further discussion of the agreement metrics.

\begin{table}[t]
    \centering
    \small
    \setlength{\tabcolsep}{3pt}
    \renewcommand{\arraystretch}{0.95}
    \caption{Independent human re-evaluation.
\textbf{Appro.}, \textbf{Relev.}, and \textbf{Pers.} denote appropriateness, relevance, and personalized response quality, respectively.}
    \label{tab:human-reevaluation}
    \resizebox{\columnwidth}{!}{%
    \begin{tabular}{lcccc}
        \toprule
        Metric & Appro. & Relev. & Pers. & \textbf{Overall} \\
        \midrule
        Positive rate (\%) & 91.50 & 91.00 & 88.17 & \textbf{90.22} \\
        Pairwise agreement (\%) & 84.68 & 83.54 & 79.25 & \textbf{82.49} \\
        Gwet's AC1 & 0.8186 & 0.8032 & 0.7377 & \textbf{0.7874} \\
        Fleiss' $\kappa$ & 0.0154 & $-0.0046$ & 0.0054 & \textbf{0.0076} \\
        \bottomrule
    \end{tabular}%
    }
\end{table}

\section{RootMem Framework}

\begin{figure}[!t]
    \centering
    \includegraphics[
        width=\columnwidth,
        trim=0.7cm 0.5cm 0.5cm 0cm,
        clip
    ]{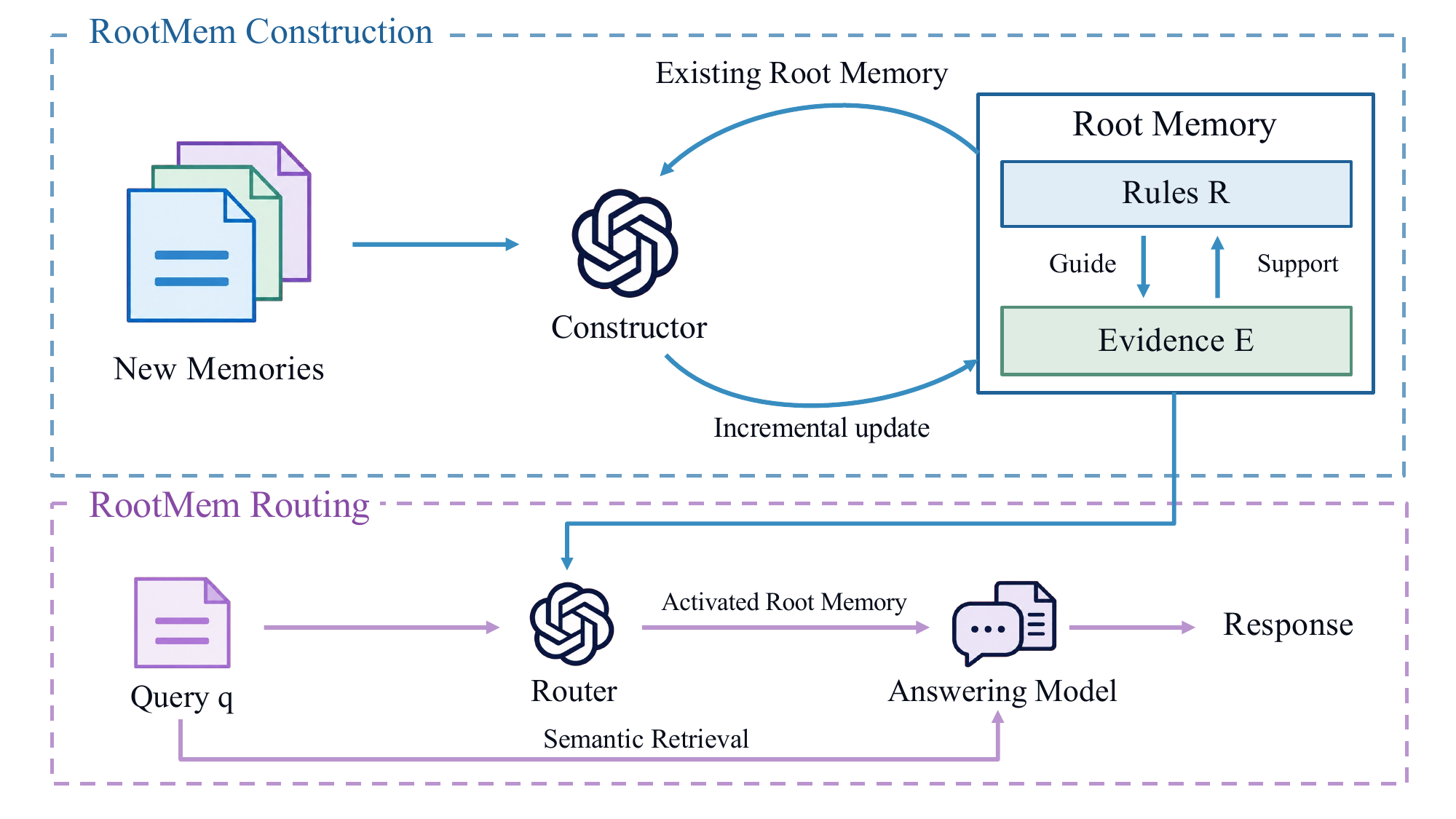}
    \caption{Overview of RootMem. Top: incremental construction of root memory units containing response-guiding rules and supporting personalized evidence. Bottom: query-conditioned routing activates relevant units, which complement semantic retrieval results to guide response generation.}
    \label{fig:main_figure_draft}
\end{figure}

To investigate whether 
organizing personalized logic into rules and supporting evidence can mitigate semantic deflection, we introduce \textbf{RootMem}, a plug-and-play framework for constructing and routing root memories. As shown in Figure~\ref{fig:main_figure_draft}, RootMem follows two stages. During construction, it distills reusable response-guiding rules and their supporting personalized evidence from user histories into root memory units, incrementally updating them as new memories arrive. During inference, an LLM-based router activates units relevant to the current query and combines them with conventional semantic retrieval results to guide response generation.

\subsection{Root Memory Construction}

During long-term interactions, raw memory entries extracted by the system are often fragmented and lack logical consolidation. To organize these memories, RootMem organizes historical memories into structured root memory units that capture reusable personalized decision logic and its supporting evidence. We denote the set of extracted root memory units as $\mathcal{U} = \{U_1, U_2, \dots, U_K\}$, where each unit $U_i$ is formally defined as:
\begin{equation}
U_i = \langle \mathcal{R}_i, \mathcal{E}_i \rangle,
\end{equation}
where the \textbf{execution rules} $\mathcal{R}_i$ specify the activation conditions and response-guiding logic of the unit, while the \textbf{personalized logical evidence} $\mathcal{E}_i$ provides the user-specific facts and constraints that support these rules.

Specifically, $\mathcal{R}_i$ determines when and how the unit should affect future responses, while $\mathcal{E}_i$ grounds this logic with supporting facts from the user's history.
A root memory unit captures not only what facts are stored, but how these facts should constrain personalized responses.

For example, consider a case where one memory records that the user often works in a studio, while another records that the user's cat frequently stays with the user there. From these fragmented memories, RootMem derives a reusable decision rule: when future queries concern the studio environment, the model should account for the cat's presence and safety. $\mathcal{E}_i$ grounds this rule with the corresponding user-specific evidence. Thus, the root memory captures an implicit constraint jointly implied by the memories, rather than either fact alone.

To construct root memories, RootMem periodically aggregates newly collected memory entries into a batch set $\mathcal{M}_t$ and invokes an LLM-based extractor $\Phi_C$ to update the existing root memory set $\mathcal{U}^{(t-1)}$. Given the new batch, $\Phi_C$ updates existing units by enriching their supporting evidence or refining their execution rules, creates new units for emerging personalized decision logic, and merges units with overlapping constraints.
The number of root memory units $K$ changes as units are created or merged, subject to a maximum unit budget that controls computational cost. The update is defined over the root memory set:
\begin{equation}
\mathcal{U}^{(t)} = \Phi_C(\mathcal{U}^{(t-1)}, \mathcal{M}_t),
\end{equation}
where $\mathcal{U}^{(t-1)}$ and $\mathcal{U}^{(t)}$ denote the root memory sets before and after incorporating the new batch $\mathcal{M}_t$, respectively.

\begin{table*}[t!]
\centering
\small
\renewcommand{\arraystretch}{0.92}
\setlength{\aboverulesep}{0.3ex}
\setlength{\belowrulesep}{0.3ex}
\caption{Memory-level MCQ results on IMLogic. Rel. Improv. is RootMem's relative accuracy gain over each retrieval baseline. Latency is in seconds per user; dashes indicate unavailable or inapplicable values.}
\label{tab:error_analysis}
\resizebox{0.95\textwidth}{!}{
\begin{tabular}{l c c ccc ccc}
\toprule
\multirow{2}{*}{\textbf{Method}} & \multirow{2}{*}{\textbf{\shortstack{Acc. (\%)}}} & \multirow{2}{*}{\textbf{\shortstack{Rel.\\Improv. (\%)}}} & \multicolumn{3}{c}{\textbf{Trap Rates (\%)}} & \multicolumn{3}{c}{\textbf{Latency (s)}} \\
\cmidrule(lr){4-6} \cmidrule(lr){7-9}
 & & & \textbf{Preference} & \textbf{Generic} &  \textbf{Fabrication} & \textbf{Consol.} & \textbf{Retrieval} & \textbf{Total} \\
\midrule
\rowcolor{gray!10} \multicolumn{9}{l}{\textbf{LLM with Full Memory}} \\
GPT-5.1 & 70.35 & -- & 18.46 & 11.10 & 0.09 & -- & -- & -- \\
Qwen3-Max & 75.54 & -- & 20.98 & 2.84 & 0.63 & -- & -- & -- \\
\midrule
\rowcolor{gray!10} \multicolumn{9}{l}{\textbf{Lexical Retrieval}} \\
BM25 & 39.62 & +39.40 & 34.79 & 25.45 & 0.14 & -- & -- & -- \\
\midrule
\rowcolor{gray!10} \multicolumn{9}{l}{\textbf{Semantic Retrieval}} \\
all-MiniLM-L6-v2 & 33.98 & +62.54 & 38.49 & 27.44 & 0.09 & -- & -- & -- \\
text-embedding-3-small & 40.93 & +34.94 & 36.73 & 22.25 & 0.09 & -- & -- & -- \\
Qwen3-Embedding-4B & 39.98 & +38.14 & 35.51 & 24.37 & 0.14 & -- & -- & -- \\
\midrule
\rowcolor{gray!10} \multicolumn{9}{l}{\textbf{Graph Retrieval}} \\
Hierarchical Cluster-Tree & 25.18 & +119.34 & 5.37 & 69.27 & 0.18 & 179.34 & 274.85 & 454.19 \\
Entity-Linked Graph & 41.83 & +32.03 & 20.71 & 37.32 & 0.14 & 6043.87 & 99.23 & 6143.10 \\
Similarity-Linked Graph & 38.99 & +41.65 & 38.13 & 22.74 & 0.14 & 4055.00 & 62.63 & 4117.63 \\
\midrule
\rowcolor{gray!10} \multicolumn{9}{l}{\textbf{Hybrid Retrieval}} \\
Reranker & 43.14 & +28.03 & 36.55 & 20.17 & 0.14 & -- & 1322.12 & 1322.12 \\
Query Reconstruction + Reranker & 43.41 & +27.23 & 36.28 & 20.17 & 0.14 & -- & 1762.14 & 1762.14 \\
\rowcolor{green!10} \textbf{RootMem (Ours)} & \textbf{55.23} & -- & 28.84 & 15.79 & 0.14 & 510.32 & 144.21 & 654.53 \\
\bottomrule
\end{tabular}
}
\end{table*}

\subsection{Root Memory Routing}

During online inference, RootMem employs a root memory routing mechanism to access logical connections that may not be captured by semantic retrieval alone. The router evaluates the execution rules $\mathcal{R}_i$ of each root memory unit to identify the units relevant to the current query.

Formally, given a user query $q$, the Root Memory Router $\Phi_R$ determines whether to activate each root memory unit based on its execution rules $\mathcal{R}_i$:
\begin{equation}
    \mathcal{U}_{\mathrm{act}} = \{ U_i \in \mathcal{U} \mid \Phi_R(q, \mathcal{R}_i) = 1 \},
\end{equation}
where $\Phi_R(q, \mathcal{R}_i) \in \{0, 1\}$ indicates whether the router deems the $i$-th root memory unit relevant to the current query.

RootMem organizes each activated unit into a unified root memory block $[\mathcal{R}_i, \mathcal{E}_i]$, which is then concatenated with the memory entries $\mathcal{M}_{\mathrm{sem}}$ retrieved by standard semantic retrieval to form a logic-guided context $C$:
\begin{equation}
C = \mathrm{Conc.}\left(
\{[\mathcal{R}_i, \mathcal{E}_i] \mid U_i \in \mathcal{U}_{\mathrm{act}}\},
\mathcal{M}_{\mathrm{sem}}
\right),
\end{equation}
where $\mathrm{Conc.}(\cdot)$ denotes the concatenation function that integrates the activated root memory blocks and the retrieved memory entries into a unified textual context.
This context augmentation complements semantic recall by injecting relevant decision-guiding logic and its supporting personalized evidence, thereby providing a more reliable basis for response generation.

\section{Experiments}

We conduct experiments on IMLogic to investigate two research questions:
(1) How effectively do existing retrieval methods and memory systems handle semantic--logical divergence, and what failure patterns emerge?
(2) To what extent can organizing personalized logic into rules and supporting evidence mitigate these failures, and what challenges remain?

Following the experimental setup (Section~\ref{sec:exp_setup}), we evaluate retrieval methods and full-memory references at the memory level (Section~\ref{sec:memory_level_results}) and memory systems at the conversation level (Section~\ref{sec:conversation_level_results}), using RootMem to explore the potential of structured personalized logic organization. We further investigate semantic deflection by analyzing the relationship between memory retrieval and diagnostic response outcomes (Section~\ref{sec:diagnosing_semantic_deflection}), and examine open-ended answering and cross-judge consistency (Section~\ref{sec:openended_results}). Additional analyses cover RootMem ablations (Appendix~\ref{sec:ablation_analysis}), sensitivity to the root memory unit budget and construction batch size (Appendix~\ref{sec:sensitivity_analysis}), and cross-backbone comparisons (Appendix~\ref{sec:backbone_analysis}).

\subsection{Experimental Setup}
\label{sec:exp_setup}

\subsubsection{Memory-Level Setup}
At the memory level, we evaluate lexical, semantic, graph-based, and hybrid retrieval methods, as well as RootMem, using pre-annotated memory entries. All retrieval-based methods use top-30 retrieval (including activated root memories for RootMem) and GPT-5.1~\citep{gpt-5.1} to answer MCQs based on the retrieved context. This setting focuses on memory retrieval and utilization without introducing variations from dialogue-level memory extraction. We additionally evaluate GPT-5.1~\citep{gpt-5.1} and Qwen3-Max~\citep{qwen2025max} with the complete memory bank as empirical capability references under full information access. GPT-5.1 matches the answering model, enabling a more direct comparison.

\subsubsection{Conversation-Level Setup}
At the conversation level, systems process raw dialogue histories end to end. We evaluate \textit{A-Mem}~\citep{xu2025amem}, \textit{Mem0} and \textit{Mem0G}~\citep{chhikara2025mem0}, \textit{TACITREE}~\citep{li2025implexconv}, \textit{MemOS}~\citep{li2025memos}, \textit{LightMem}~\citep{fang2025lightmem}, \textit{EverMemOS}~\citep{hu2026evermemos}, and \textit{T-Mem}~\citep{guo2026t}, with GPT-4o-mini~\citep{openai2024gpt4omini} as the shared backbone and GPT-5.1 as the shared MCQ answering model. All methods use the same top-30 retrieval setting. We further augment MemOS, LightMem and EverMemOS with RootMem to examine whether structured personalized logic can complement existing memory systems.

\subsubsection{Evaluation Protocols}
The main comparisons and ablation studies use full IMLogic. 
Due to computational cost, sensitivity analysis, open-ended evaluation, and cross-backbone comparisons each use a randomly sampled five-user subset. 
For open-ended evaluation, GPT-4o-mini serves as the memory-system backbone, while GPT-5.1~\citep{gpt-5.1} serves as the answering model, with DeepSeek-flash~\citep{deepseek2026apidocs} and Qwen3.7-Max~\citep{qwen2026max37} independently judging the generated responses against the logical requirements of the references. For cross-backbone evaluation, we vary the memory system LLM while keeping GPT-5.1 fixed as the MCQ answering model. Further details on baselines, model configurations, ablations, and parameter settings are provided in Appendix~\ref{sec:baseline_details}.

\begin{table*}[t!]
\centering
\small
\renewcommand{\arraystretch}{0.92}
\setlength{\aboverulesep}{0.3ex}
\setlength{\belowrulesep}{0.3ex}
\caption{Conversation-level MCQ results on IMLogic. Parentheses indicate relative changes from the corresponding system without RootMem.}
\label{tab:end_to_end}
\resizebox{0.9\textwidth}{!}{
\begin{tabular}{l c c w{c}{2.3cm} w{c}{2.3cm} w{c}{2.3cm}} 
\toprule
\multirow{2}{*}{\textbf{System}} & \multirow{2}{*}{\textbf{Retrieval Type}} & \multirow{2}{*}{\textbf{\shortstack{Acc. (\%)}}} & \multicolumn{3}{c}{\textbf{Trap Rates (\%)}} \\
\cmidrule(lr){4-6}
& & & \textbf{Preference} & \textbf{Generic} & \textbf{Fabrication} \\
\midrule
A-Mem & Graph & 37.32 & 29.60 & 32.90 & 0.18 \\
Mem0 & Semantic & 38.18 & 22.11 & 39.53 & 0.18 \\
Mem0G & Graph & 38.09 & 23.10 & 38.67 & 0.14 \\
TACITREE & Graph & 30.46 & 14.12 & 55.37 & 0.05 \\
T-Mem & Hybrid & 44.18 & 23.92 & 31.86 & 0.05 \\
\midrule
MemOS & Hybrid & 39.44 & 28.11 & 32.31 & 0.14 \\
\rowcolor{green!10} \quad + RootMem & Hybrid & \textbf{51.04} (+29.41\%) & 24.10 (-14.26\%) & 24.77 (-23.34\%) & 0.09 (-35.71\%) \\
\midrule
LightMem & Semantic & 37.55 & 29.69 & 32.67 & 0.09 \\
\rowcolor{green!10} \quad + RootMem & Hybrid & \textbf{50.23} (+33.77\%) & 25.54 (-13.98\%) & 24.14 (-26.11\%) & 0.09 (+0.00\%) \\
\midrule
EverMemOS & Hybrid & 43.50 & 27.30 & 29.11 & 0.09 \\
\rowcolor{green!10} \quad + RootMem & Hybrid & \textbf{50.41} (+15.89\%) & 23.51 (-13.88\%) & 25.99 (-10.72\%) & 0.09 (+0.00\%) \\
\bottomrule
\end{tabular}
}
\end{table*}

\subsubsection{Metrics}

We report MCQ accuracy and the rates of three diagnostic response categories: Preference Trap, Fabrication Trap, and Generic Trap, which respectively probe semantic deflection, hallucination, and lack of personalization. Some methods consolidate or synthesize memories, so retrieved units may not correspond one-to-one to the original memory entries. We therefore use answer-level metrics for cross-method comparisons, while separately analyzing semantic and logical memory recall using Recall@30 in Section~\ref{sec:diagnosing_semantic_deflection}. Open-ended accuracy measures the proportion of responses accepted by each judge, while judge agreement measures the proportion receiving consistent correctness judgments. Latency is reported in seconds per user, separating offline consolidation from online retrieval.

\subsection{Memory-Level Evaluation}
\label{sec:memory_level_results}

Table~\ref{tab:error_analysis} reveals three key findings:
\begin{enumerate}[label=\textbf{(\alph*)}, leftmargin=*, itemsep=3pt, topsep=3pt]

\item \textbf{Existing retrieval methods remain vulnerable to semantic deflection.}
The strongest retrieval baseline achieves only 43.41\% accuracy. Graph-based retrieval does not consistently outperform lexical or semantic retrieval, while query reconstruction offers only marginal gains over reranking. Moreover, lexical, semantic, and hybrid methods produce Preference Trap responses on 34.79--38.49\% of questions, frequently favoring apparent preferences in $m_s$ while overlooking the decisive logic of $m_l$. These findings highlight the difficulty of effectively utilizing logically critical memories with existing retrieval strategies.

\item \textbf{Full-memory access provides a capability reference, but utilizing logical memories remains challenging.}
With access to the complete memory bank, both GPT-5.1 and Qwen3-Max substantially outperform retrieval-based methods and provide an empirical capability reference for IMLogic. Nevertheless, 18.46\% of GPT-5.1 responses and 20.98\% of Qwen3-Max responses still fall into the Preference Trap category. These results suggest that effectively using logical memories remains challenging even with full-memory access, as performance also depends on the answering model's long-context understanding and reasoning capabilities.

\item \textbf{Organizing personalized logic into rules and supporting evidence offers a promising solution.}
RootMem achieves 55.23\% accuracy, a 27.23\% relative improvement over the strongest retrieval baseline, while reducing Preference Trap rate from 36.28\% to 28.84\%. These gains are accompanied by competitive total latency and low online retrieval overhead, suggesting that structured logical memory organization can improve effectiveness while remaining computationally efficient. Nevertheless, substantial errors remain, indicating that reliably utilizing implicit personalized logic remains challenging.

\end{enumerate}

\subsection{Conversation-Level Evaluation}
\label{sec:conversation_level_results}

Table~\ref{tab:end_to_end} reveals two key findings:
\begin{enumerate}[label=\textbf{(\alph*)}, leftmargin=*, itemsep=3pt, topsep=3pt]
\item \textbf{End-to-end memory systems struggle with implicit logical memory retrieval.}
Baseline accuracies range from 30.46\% to 44.18\% across memory systems with different retrieval strategies. Among the baselines, hybrid systems achieve higher accuracies overall, while graph-based systems show no consistent advantage over semantic systems. Notably, T-Mem, which explicitly supports associative recall through future-oriented triggers, achieves the highest accuracy, but 23.92\% of its answers still fall into the Preference Trap, and 31.86\% into the Generic Trap category. Fabrication Trap responses remain rare in this MCQ setting.

\item \textbf{Organizing personalized logic improves performance across memory systems.}
Augmenting MemOS, LightMem, and EverMemOS with RootMem increases accuracy to 51.04\%, 50.23\%, and 50.41\%, respectively, averaging a 26.36\% relative gain. On average, Preference Trap rate and Generic Trap rate decrease by 14.04\% and 20.06\% in relative terms, respectively. These reductions suggest that RootMem helps systems select personalized responses that account for relevant logical constraints. Nevertheless, Preference Trap and Generic Trap responses still account for 23.51--25.54\% and 24.14--25.99\% of answers, respectively, leaving substantial room for further improvement on IMLogic.
\end{enumerate}

\subsection{Diagnosing Semantic Deflection}
\label{sec:diagnosing_semantic_deflection}
\begin{table}[t]
\centering
\small
\setlength{\tabcolsep}{2pt}
\renewcommand{\arraystretch}{1.04}
\caption{Answer outcomes by retrieved target memories. Accuracy and trap rates are percentages within each row.}
\label{tab:diagnosing_semantic_deflection}
\begin{tabular*}{\columnwidth}{@{\extracolsep{\fill}}lccccc@{}}
\toprule
\multirow[c]{2}{*}[-2.5pt]{\shortstack[l]{\textbf{Retrieved}\\\textbf{Memories}}} &
\multirow[c]{2}{*}[-2.5pt]{\textbf{\# QA}} &
\multirow[c]{2}{*}[-2.5pt]{\textbf{\shortstack{Acc. (\%)}}} &
\multicolumn{3}{c}{\textbf{Trap Rates (\%)}} \\
\cmidrule(lr){4-6}
 & & & \textbf{Preference} & \textbf{Generic} & \textbf{Fabrication} \\
\midrule
$m_s+m_l$ & 470   & 60.85 & 28.94 & 10.21 & 0.00 \\
$m_s$ only & 1,362 & 22.69 & 49.05 & 28.12 & 0.15 \\
$m_l$ only & 126   & 75.40 & 4.76  & 19.84 & 0.00 \\
Neither    & 258   & 24.42 & 16.67 & 58.91 & 0.00 \\
\midrule
Overall    & 2,216 & 33.98 & 38.49 & 27.44 & 0.09 \\
\bottomrule
\end{tabular*}
\end{table}

To further examine IMLogic's ability to diagnose semantic deflection, we analyze how the retrieval of semantic memory $m_s$ and logical memory $m_l$ relates to response outcomes. We examine the top-30 results of all-MiniLM-L6-v2, which directly retrieves original memory entries and allows their recall to be measured. Table~\ref{tab:diagnosing_semantic_deflection} reports the results across IMLogic, revealing three findings:

\begin{enumerate}[label=\textbf{(\alph*)}, leftmargin=*, itemsep=3pt, topsep=3pt]

\item \textbf{Semantic--Logical Recall Gap.}
Recall@30 is 82.67\% for $m_s$ but only 26.90\% for $m_l$. In 61.46\% of questions, $m_s$ is retrieved without $m_l$, accounting for 78.31\% of all Preference Trap responses. This demonstrates a prevalent mismatch between semantic relevance and logical importance, closely associated with Preference Trap errors.

\item \textbf{Logical Memory Availability and Answer Accuracy.}
Accuracy reaches 63.93\% when $m_l$ is retrieved, but only 22.96\% otherwise, below the 25\% random-choice baseline. Correct answers without the annotated $m_l$ may rely on alternative supporting memories or occur by chance. Nevertheless, 36.07\% of responses remain incorrect even with $m_l$. Consistent with results in Section~\ref{sec:memory_level_results}, this highlights the importance of reasoning and contextual understanding beyond retrieval.

\item \textbf{Semantic Memory and Response Failure Patterns.}
\textit{(i) When $m_l$ is retrieved,} Preference Trap rate is 28.94\% with $m_s$, compared with 4.76\% without it. This suggests that semantically similar memories are associated with failures to utilize available logical evidence. \textit{(ii) When $m_l$ is not retrieved,} the proportion of Preference Trap responses among incorrect answers increases from 22.05\% without $m_s$ to 63.44\% with it. The dominant error type shifts from Generic Trap to Preference Trap, while accuracy remains nearly unchanged (24.42\% and 22.69\%, respectively). Thus, the presence of $m_s$ is associated with a shift from generic errors toward superficially personalized responses that overlook important logic.

\end{enumerate}

\begin{table}[t]
\centering
\small
\setlength{\tabcolsep}{3pt}
\renewcommand{\arraystretch}{1.0}
\setlength{\aboverulesep}{0.3ex}
\setlength{\belowrulesep}{0.3ex}
\caption{Open-ended results at the conversation level on sampled five-user subset, evaluated by two LLM judges. Parentheses indicate relative accuracy gains from RootMem.}
\label{tab:robustness}
\label{tab:openended_generalization}
\label{tab:judge_consistency}
\renewcommand{\tabularxcolumn}[1]{m{#1}}
\begin{tabularx}{\columnwidth}{@{}l*{3}{>{\centering\arraybackslash}X}@{}}
\toprule
\multirow[c]{2}{*}{\textbf{System}} &
\multicolumn{2}{c}{\textbf{Accuracy by judge (\%)}} &
\multirow[c]{2}{*}{\shortstack{\textbf{Judge}\\\textbf{agreement (\%)}}} \\
\cmidrule(lr){2-3}
& \textbf{DeepSeek-flash} & \textbf{Qwen3.7-Max} & \\
\midrule
Mem0 & 13.85 & 15.19 & 89.42 \\
Mem0G & 15.00 & 16.92 & 92.69 \\
\midrule
LightMem & 17.69 & 18.27 & 93.27 \\
\rowcolor{green!10} \quad + RootMem & \mbox{\textbf{26.92} {\footnotesize (+52.18\%)}} & \mbox{\textbf{27.69} {\footnotesize (+51.56\%)}} & 89.23 \\
\midrule
EverMemOS & 17.50 & 20.38 & 90.58 \\
\rowcolor{green!10} \quad + RootMem & \mbox{\textbf{20.19} {\footnotesize (+15.37\%)}} & \mbox{\textbf{23.08} {\footnotesize (+13.25\%)}} & 90.96 \\
\bottomrule
\end{tabularx}
\end{table}

\subsection{Open-Ended Evaluation}
\label{sec:openended_results}


Open-ended queries require models to identify and apply personalized constraints without predefined answer options. As shown in Table~\ref{tab:openended_generalization}, baseline accuracies remain around 20\% or below, while RootMem improves both LightMem and EverMemOS. LightMem shows larger relative gains than EverMemOS under both judges. Nevertheless, even augmented systems achieve less than 30\% accuracy, highlighting the difficulty of reliably applying implicit personalized logic in free-form responses. These results suggest that structured logical organization remains beneficial, but its effectiveness is limited in this more realistic and demanding setting.

\textbf{Cross-Judge Consistency.}
To assess whether the open-ended evaluation results depend on a particular judge, we evaluate the same generated responses using DeepSeek-flash and Qwen3.7-Max. As shown in Table~\ref{tab:judge_consistency}, the two judges agree on 89.23--93.27\% of responses and consistently report improvements from RootMem augmentation. These results suggest that the observed gains are not specific to a single judge's scoring preferences, reducing concerns about judge-specific bias.

\section{Conclusion}

In this paper, we investigate semantic deflection in personalized LLM memory systems and introduce IMLogic, a benchmark for implicit logical memory retrieval in long dialogues. With diagnostic response options and evaluation at both memory and conversation levels, IMLogic supports systematic analysis of semantic deflection, hallucination, and lack of personalization. Extensive experiments reveal persistent challenges across retrieval methods and memory systems, including substantial preference-driven errors even under full-memory access. Further analysis reveals a substantial semantic–logical recall gap and persistent errors even when logical memories are retrieved. To address this challenge, we introduce root memory and instantiate it through RootMem, demonstrating the potential of organizing personalized logic into rules and supporting evidence to improve memory utilization. Nevertheless, considerable room for improvement remains, particularly in open-ended generation. Future work will explore more effective ways to extract, consolidate, organize, and retrieve reusable personalized logic, as well as the potential of parametric memory for representing and utilizing such logic.



\bibliographystyle{ACM-Reference-Format}
\bibliography{main}


\appendix
\section{Analysis of Logical Associations}
\label{sec:logic}

We summarize the mined logical associations into seven categories through post hoc analysis to characterize the diversity of IMLogic. The categories describe how a logical memory $m_l$ changes the suitability, feasibility, or priority of a response suggested by the semantically relevant background in $m_s$. The two memories can both be factually valid: their association concerns how personal context should shape the current response. These categories summarize the mined associations rather than prescribe rules for data construction. Due to space limit, detailed definitions and examples are available in our repository.

\begin{itemize}[leftmargin=*, itemsep=2pt]

\item \textbf{State Constraint}

\item \textbf{Motivational Orientation}

\item \textbf{Identity Shift}

\item \textbf{Goal Alignment}

\item \textbf{Contextual Fit}

\item \textbf{Resource Constraint}

\item \textbf{Others}

\end{itemize}

Together, these categories characterize the personal conditions that can remain hidden behind a semantically relevant memory. They make IMLogic's central distinction explicit: topical relevance identifies useful background, while logical relevance determines how that background should affect the current response.

\begin{table*}[t]
\centering
\small
\setlength{\tabcolsep}{4pt}
\begin{threeparttable}
\caption{Abridged example from the human evaluation form used for dataset-quality assessment. The query, memories, and response options are shortened for presentation.}
\label{tab:human-eval-sample}
\begin{tabularx}{\textwidth}{>{\raggedright\arraybackslash\bfseries}p{2.4cm} X}
\toprule
Query &
Where should I go to avoid the chaos of my last vacation? \\
\midrule
Semantic Memory &
Donald disliked his last crowded city trip due to the lack of peace and personal space. \\
Logical Memory &
Donald wants more cultural experiences when traveling. \\
\midrule
Choices &
\textbf{Correct Response:} Visit a quiet historic town with museums and cultural sites, offering both tranquility and cultural experiences. \newline
\textbf{Preference Trap:} Choose a remote cabin for complete isolation and peace. \newline
\textbf{Fabrication Trap:} Try the ``Silent City'' resort, designed for your medical sensitivity to noise and crowds. \newline
\textbf{Generic Trap:} Consider a beach resort or mountain retreat for a relaxing getaway. \\
\midrule
Evaluation Criteria &
\textbf{Appropriateness}: The query is well-formed and natural. \newline
\textbf{Relevance}: The query is consistent with the provided memory context. \newline
\textbf{Personalized Response}: The first response is clearly more personalized and better grounded in memory than the alternatives. \\
Scale &
Annotators rated each dimension on a 4-point scale: \newline
\textbf{1} = Disagree \qquad \textbf{2} = Somewhat disagree \newline
\textbf{3} = Somewhat agree \qquad \textbf{4} = Agree \\
\bottomrule
\end{tabularx}
\end{threeparttable}
\end{table*}

%
\section{Human Evaluation on Dataset Quality}
\label{sec:human_evaluation}
We conduct an independent re-evaluation on 30 QA instances randomly sampled from different users. Twenty annotators recruited through CloudResearch evaluate every instance along the same three dimensions used during construction-time expert validation: \textbf{(1) Appropriateness}, whether the query is well-formed and natural; \textbf{(2) Relevance}, whether the query is consistent with the semantic and logical memories and constitutes a plausible user request; and \textbf{(3) Personalized Response Quality}, whether the designated correct response is more personalized and better grounded in the provided memories than the alternatives. Table~\ref{tab:human-eval-sample} presents an example evaluation item.

The annotators are primarily from North America. Their ages range from 24 to 78 years with a median of 36.5, and 15 of the 20 hold a bachelor's degree or above. Each annotator evaluates all 30 instances using a four-point scale, with a median completion time of 25.0 minutes. We code Agree and Somewhat Agree as positive, and Somewhat Disagree and Disagree as negative. Each dimension contains 600 binary ratings; overall statistics pool all 1,800 ratings.

\textbf{Agreement Metrics.} Table~\ref{tab:human-reevaluation} reports high positive rates and raw agreement across the three evaluation dimensions. The near-zero Fleiss' $\kappa$~\citep{fleiss1971agreement} indicates little agreement beyond that expected from the strongly skewed positive labels. Gwet's AC1~\citep{gwet2008agreement} uses a different adjustment for chance agreement and is less sensitive to this imbalance. We therefore report both coefficients alongside raw agreement and positive rates to provide a fuller account of the annotation results.

\section{Baselines and Implementation Details}
\label{sec:baseline_details}
Section~\ref{sec:exp_setup} defines the evaluation inputs, sample coverage, and metrics. Here we detail the baseline mechanisms, model assignments, and construction configurations used to reproduce the comparisons.
\subsection{Memory-Level Experiments}
\subsubsection{Evaluated Baselines}
To comprehensively evaluate implicit logical memory retrieval performance, we categorize the selected baselines into four groups:

\begin{itemize}[leftmargin=*, nosep]
\item \textbf{Lexical Retrieval}: We employ \textit{BM25}~\citep{robertson2009probabilistic} as the standard sparse retrieval baseline. 

\item \textbf{Semantic Retrieval}: We utilize three widely adopted dense retrievers, including \textit{all-MiniLM-L6-v2}~\citep{all-MiniLM-L6-v2}, \textit{text-embedding-3-small}~\citep{text-embedding-3-small}, and \textit{Qwen3-Embedding-4B}~\citep{zhang2025qwen3}. 

\item \textbf{Graph Retrieval}: We select three representative graph-based retrieval baselines. The \textit{Hierarchical Cluster-Tree}, based on TACITREE~\citep{li2025implexconv}, structures memory history into multiple levels of summarization to enable top-down hierarchical retrieval. The \textit{Entity-Linked Graph}, built upon Mem0G~\citep{chhikara2025mem0}, extracts key entities and captures their complex relationships to support associative retrieval. The \textit{Similarity-Linked Graph}, derived from A-Mem~\citep{xu2025amem}, dynamically establishes links between memory nodes based on semantic similarity.

\item \textbf{Hybrid Retrieval}: Based on the agentic search module from EverMemOS~\citep{hu2026evermemos}, we implement two hybrid retrieval baselines. The \textit{Reranker} method performs dual-path retrieval using \textit{BM25}~\citep{robertson2009probabilistic}  and \textit{Qwen3-Embedding-4B}~\citep{zhang2025qwen3}, followed by reranking with \textit{Qwen3-Reranker-4B}~\citep{zhang2025qwen3}. The \textit{Query Reconstruction + Reranker} method first rewrites the original query into 2-3 sub-queries, executes the aforementioned multi-path retrieval, and performs a final reranking.
\end{itemize}

\subsubsection{Implementation Details}
For the main memory-level comparison, retrieval methods that require an LLM use \texttt{GPT-4o}~\citep{gpt-4o} as their backbone and the retrieval cutoff is fixed at 30 for all retrieval-based methods. The answering model for MCQ is \texttt{GPT-5.1}. RootMem also uses GPT-4o for root memory construction and routing, with \textit{all-MiniLM-L6-v2} as the dense retriever for semantic compensation. We set the maximum number of \yxnew{root memory} units to 10 and the \yxnew{root memory} construction batch size to 20.

\subsection{Conversation-Level Experiments}

\subsubsection{Evaluated Baselines}
To examine implicit logical memory retrieval in complete memory systems, we evaluate the following baselines:
\begin{itemize}[leftmargin=*, nosep]
    \item \textbf{A-Mem}~\citep{xu2025amem}: Draws inspiration from the Zettelkasten method to build interconnected knowledge networks through dynamic indexing and linking.
    \item \textbf{Mem0}~\citep{chhikara2025mem0}: A production-ready system that manages long-term memory by dynamically extracting and consolidating salient information.
    \item \textbf{Mem0G}~\citep{chhikara2025mem0}: An enhanced variant of Mem0 that incorporates graph representations to capture complex relational structures.
    \item \textbf{TACITREE}~\citep{li2025implexconv}: Structures long conversation histories into hierarchical summarization trees for multi-level retrieval.
    \item \textbf{MemOS}~\citep{li2025memos}: Adopts an OS-inspired architecture, elevating memory to a first-class operational resource with structured management.
    \item \textbf{LightMem}~\citep{fang2025lightmem}: A lightweight long-term memory framework optimized for large-scale retrieval efficiency.
    \item \textbf{EverMemOS}~\citep{hu2026evermemos}: Implements an engram-inspired lifecycle for comprehensive memory management.
    \item \textbf{T-Mem}~\citep{guo2026t}: Organizes conversational memories into scenes and items, using future-oriented triggers to support associative retrieval.
\end{itemize}

\subsubsection{Implementation Details}
To examine the role of structured personalized logic across memory systems, we integrate RootMem into three representative baselines (MemOS, LightMem, and EverMemOS) and directly process the memories produced by each system. Since these memory systems extract substantially more memories than the original human-annotated entries, we increase the root memory construction batch size to 60. To control computational cost, we use \texttt{GPT-4o-mini} as the backbone for all evaluated systems, including RootMem construction and routing, while using \texttt{GPT-5.1} as the answering model. Other shared configurations follow the memory-level setup.

\subsection{Open-Ended Evaluation Protocol}
\label{sec:extended_protocol}

We conduct open-ended evaluation on a randomly sampled five-user subset, using GPT-4o-mini as the memory-system backbone and GPT-5.1 as the answering model. Responses are generated without candidate options and independently evaluated by DeepSeek-flash and Qwen3.7-Max based on whether they incorporate the underlying logic of $m_l$. We report accuracy under each judge and their agreement, with relative improvements computed separately for each judge. Generation and judging prompts are provided in our repository.

\subsection{Additional Analysis Protocols}
\label{sec:ablation_protocol}

\noindent\textbf{Component Ablations.}
We conduct ablations under the memory-level MCQ setting, primarily modifying the root memory routing stage. Without routing, all root memory units are activated. Removing execution rules retains personalized logical evidence, while removing evidence retains execution rules. Semantic Only uses all-MiniLM-L6-v2 without root memory augmentation.

\noindent\textbf{Sensitivity Analysis.}
We evaluate sensitivity on a randomly sampled five-user subset under the memory-level MCQ setting. We vary the maximum root memory unit budget over $\{5,10,20\}$ with batch size fixed at 20, and the construction batch size over $\{10,20,40\}$ with the unit budget fixed at 10. The unit budget specifies an upper limit rather than a required number of units.

\noindent\textbf{Cross-Backbone Evaluation.}
We evaluate GPT-4o-mini, Qwen3-30B-A3B-Instruct-2507~\cite{qwen2025instruct2507}, and GLM-4.6~\cite{zai2025glm-4-6} as memory-system backbones on a randomly sampled five-user subset. RootMem construction and routing use the corresponding backbone, while GPT-5.1 remains fixed as the MCQ answering model. Predictions are scored against the correct options without an LLM judge. Relative gains compare each augmented system with its corresponding baseline under the same backbone.




\section{Additional Experimental Analyses}
\label{sec:additional_experiments}

\subsection{Ablation Study}
\label{sec:ablation_analysis}
\begin{figure}[H]
    \centering
    \includegraphics[width=\columnwidth,trim=0pt 4pt 0pt 0pt,clip]{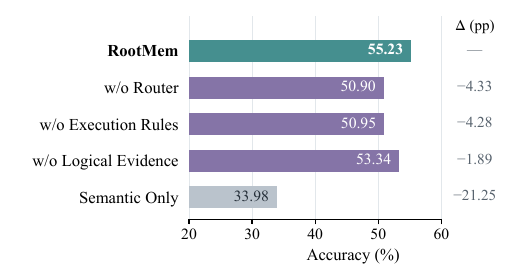}
    \caption{Ablation study on IMLogic. $\Delta$ denotes the accuracy change from RootMem in percentage points.}
    \label{fig:ablation}
    \Description{Horizontal bars compare RootMem at 55.23 percent with variants without the router at 50.90, execution rules at 50.95, or personalized logical evidence at 53.34, and semantic retrieval alone at 33.98. Their accuracy drops are 4.33, 4.28, 1.89, and 21.25 percentage points, respectively.}
\end{figure}


Figure~\ref{fig:ablation} shows that removing execution rules, personalized evidence, or query-conditioned routing consistently degrades performance. Execution rules provide guidance on when and how personal constraints should shape responses, while personalized evidence grounds this guidance in specific information. The larger drops from removing rules or routing highlight the importance of structured decision guidance and selective memory activation.

Notably, activating all root memory units without routing performs worse under the same budget, highlighting the importance of selective activation. Even without execution rules, RootMem retains 50.95\% accuracy, substantially outperforming semantic-only retrieval, suggesting that organized personalized evidence also contributes to the improvement. Together, these results highlight the complementary value of structured evidence, explicit rules, and selective activation.

\subsection{Sensitivity Analysis}
\label{sec:sensitivity_analysis}

\begin{figure}[H]
    \centering
    \includegraphics[width=\columnwidth,trim=0pt 5pt 0pt 0pt,clip]{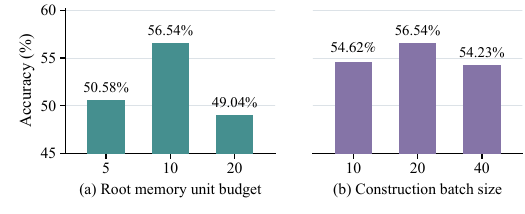}
    \caption{Sensitivity to (a) the root memory unit budget and (b) construction batch size on the evaluation subset.}
    \label{fig:sensitivity}
    \Description{Two bar charts show MCQ accuracy. With root memory unit budgets of 5, 10, and 20, accuracy is 50.58, 56.54, and 49.04 percent. With offline construction batch sizes of 10, 20, and 40, accuracy is 54.75, 56.54, and 54.23 percent.}
\end{figure}

\textbf{(a) A larger root memory unit budget does not guarantee better performance.}
Figure~\ref{fig:sensitivity}(a) exhibits a non-monotonic trend, with the best accuracy achieved at a unit budget of 10. This suggests that increasing the capacity for storing logical information does not guarantee more effective utilization.
\textbf{(b) Performance is relatively stable across construction batch sizes.}
As shown in Figure~\ref{fig:sensitivity}(b), accuracy varies only modestly across the tested batch sizes, indicating relatively low sensitivity to construction granularity within this range.

\subsection{Effect of LLM Backbones}
\label{sec:backbone_analysis}
\begin{table}[H]
\centering
\footnotesize
\setlength{\tabcolsep}{2pt}
\renewcommand{\arraystretch}{1.12}
\caption{MCQ Accuracy across memory-system LLM backbones. Parentheses indicate relative Accuracy gains from RootMem.}
\label{tab:backbone_generalization}
\renewcommand{\tabularxcolumn}[1]{m{#1}}
\begin{tabularx}{\columnwidth}{@{}l*{3}{>{\centering\arraybackslash}X}@{}}
\toprule
\multirow[c]{2}{*}{\textbf{System}} & \multicolumn{3}{c}{\textbf{MCQ Accuracy (\%)}} \\
\cmidrule(lr){2-4}
& \textbf{\shortstack{GPT-4o-\\mini}}
& \textbf{\shortstack{Qwen3-30B-A3B\\Instruct-2507}}
& \textbf{GLM-4.6} \\
\midrule
Mem0 & 34.84 & 36.10 & 35.92 \\
Mem0G & 34.66 & 39.17 & 34.30 \\
\midrule
LightMem & 37.18 & 36.28 & 32.31 \\
\rowcolor{green!10} \quad + RootMem & \mbox{\textbf{52.53} {\footnotesize (+41.29\%)}} & \mbox{\textbf{55.60} {\footnotesize (+53.25\%)}} & \mbox{\textbf{55.42} {\footnotesize (+71.52\%)}} \\
\midrule
EverMemOS & 44.04 & 43.86 & 41.16 \\
\rowcolor{green!10} \quad + RootMem & \mbox{\textbf{50.90} {\footnotesize (+15.58\%)}} & \mbox{\textbf{57.22} {\footnotesize (+30.46\%)}} & \mbox{\textbf{59.21} {\footnotesize (+43.85\%)}} \\
\bottomrule
\end{tabularx}
\end{table}

Table~\ref{tab:backbone_generalization} shows that implicit logical memory retrieval remains challenging across the three tested memory system backbones, with all unaugmented systems below 45\% accuracy. System rankings vary while the MCQ answering model remains fixed, indicating sensitivity to the LLM used for memory processing. Integrating RootMem into LightMem and EverMemOS consistently improves accuracy across all tested backbones. Notably, the absolute gains vary across backbones, ranging from 15.35 to 23.11 percentage points for LightMem and from 6.86 to 18.05 points for EverMemOS. Both systems achieve their largest gains with GLM-4.6, where their baseline accuracies are lowest. This suggests that the magnitude of RootMem's benefit depends on the underlying memory-processing backbone. 

These findings suggest that the challenges of implicit logical memory retrieval persist across different LLM backbones, while the benefits of structured logical organization are not confined to a particular LLM backbone.

\end{document}